\documentclass[11pt]{article}

\usepackage{amsmath,amsthm,amssymb}
\usepackage{authblk}
\usepackage[top=1in,bottom=1in,left=1.3in,right=1.3in]{geometry} %
\usepackage[utf8]{inputenc}
\usepackage[T1]{fontenc}
\usepackage{hyperref}
\usepackage{url}
\usepackage{booktabs}
\usepackage{amsfonts}
\usepackage{nicefrac}
\usepackage{microtype}
\usepackage{xcolor}

\usepackage{natbib}

\definecolor{mydarkblue}{rgb}{0,0.08,0.45}
\hypersetup{ %
  colorlinks=true,
  linkcolor=mydarkblue,
  citecolor=mydarkblue,
  filecolor=mydarkblue,
  urlcolor=mydarkblue,
}

\title{\Huge Quantum Maximum Entropy Inference and Hamiltonian Learning}

\author[a,b]{Minbo Gao} %
\author[c]{Zhengfeng Ji\thanks{Corresponding author: jizhengfeng@tsinghua.edu.cn}} %
\author[c]{Fuchao Wei} %
\affil[a]{ %
  Institute of Software, Chinese Academy of Sciences, Beijing, China
}
\affil[b]{
  University of Chinese Academy of Sciences, Beijing, China
}
\affil[c]{ %
  Department of Computer Science and Technology,\protect\\
  Tsinghua University, Beijing, China
}
\usepackage{algorithmic}

\usepackage[capitalize,noabbrev]{cleveref} % after hyperref

\usepackage[english]{babel} %
\usepackage{bbm} %
\usepackage{booktabs} %
\usepackage{color} %
\usepackage{derivative} %
\usepackage{dsfont} %
\usepackage{enumitem} %
\usepackage{float} %
\usepackage[T1]{fontenc} %
\usepackage{framed} %

\usepackage{graphicx} %
\usepackage{hyphenat} %
\usepackage[utf8]{inputenc} %
\usepackage{interval} %
\usepackage{mathrsfs} %
\usepackage{mathtools} %
\usepackage{mdframed} %
\usepackage{microtype} %
\usepackage{soul} %
\usepackage{times} % charter, fourier, mathpazo, times
\usepackage{tikz} %
\usepackage{subfigure} %
\usepackage{suffix} % for *-version commands
\usepackage{xparse} %
\usepackage{xspace} %
\usepackage{url} %
\usepackage{wrapfig}

\setlist[enumerate]{nolistsep,itemsep=3pt,topsep=3pt,leftmargin=*} %
\setlist[itemize]{nolistsep,itemsep=3pt,topsep=3pt,leftmargin=2em} %

\intervalconfig{soft open fences}

\newfloat{algorithm}{hbt}{lop} %
\floatname{algorithm}{Algorithm} %
\crefname{algorithm}{Algorithm}{Algorithms} %

\newfloat{metaalgorithm}{hbt}{lop} %
\floatname{metaalgorithm}{Meta-Algorithm} %
\crefname{metaalgorithm}{Meta-Algorithm}{Meta-Algorithms} %

\theoremstyle{plain}
\newtheorem{theorem}{Theorem}[section]
\newtheorem{proposition}[theorem]{Proposition}

\theoremstyle{definition}

\theoremstyle{remark}

\newcommand{\microspace}{\mspace{.5mu}} %
\newcommand{\ket}[1]{\ensuremath{\lvert\microspace#1%
    \microspace\rangle}} %
\newcommand{\bra}[1]{\ensuremath{\langle\microspace#1%
    \microspace\rvert}} %

\newcommand{\ignore}[1]{} %
\newcommand{\ip}[2]{\ensuremath{\left\langle#1,#2\right\rangle}} %
\newcommand{\norm}[1]{\ensuremath{\left\lVert #1 \right\rVert}} %
\newcommand{\val}[1]{\ensuremath{\left\langle#1\right\rangle}} %

\newcommand{\complex}{\mathbb{C}} %
\newcommand{\real}{\mathbb{R}} %

\newcommand{\dom}[1]{\mathrm{dom}\,{#1}} %
\newcommand{\cl}[1]{\mathrm{cl}\left(#1\right)} %

\newcommand{\setft}[1]{\mathrm{#1}} %
\newcommand{\Density}{\setft{D}} %
\newcommand{\Pos}{\setft{Pos}} %
\newcommand{\Herm}{\setft{Herm}} %

\newcommand{\labelstyle}[1]{\mathrm{#1}} %
\newcommand{\QIS}{\labelstyle{QIS}} %
\newcommand{\GD}{\labelstyle{GD}} %
\newcommand{\lint}{\labelstyle{int}} %
\newcommand{\lbd}{\labelstyle{bd}} %
\newcommand{\lf}{\mathscr{L}} % Linear family
\newcommand{\ef}{\mathscr{E}} % Projection family
\newcommand\restr[2]{{% we make the whole thing an ordinary symbol
  \left.\kern-\nulldelimiterspace% automatically resize the bar with \right
  #1 % the function
  \vphantom{\big|} % pretend it's a little taller at normal size
  \right|_{#2} % this is the delimiter
  }}

\DeclareMathOperator{\tr}{tr} %
\def\I{\mathds{1}} % Id
\def\L{\mathcal{L}} % Legendre-Bregman projection
\def\X{\mathcal{X}} % Space
\begin{document}

\maketitle

\begin{abstract}
  Maximum entropy inference and learning of graphical models are pivotal tasks
  in learning theory and optimization.
  This work extends algorithms for these problems, including generalized
  iterative scaling (GIS) and gradient descent (GD), to the quantum realm.
  While the generalization, known as quantum iterative scaling (QIS), is
  straightforward, the key challenge lies in the non-commutative nature of
  quantum problem instances, rendering the convergence rate analysis
  significantly more challenging than the classical case.
  Our principal technical contribution centers on a rigorous analysis of the
  convergence rates, involving the establishment of both lower and upper bounds
  on the spectral radius of the Jacobian matrix for each iteration of these
  algorithms.
  Furthermore, we explore quasi-Newton methods to enhance the performance of QIS
  and GD\@.
  Specifically, we propose using Anderson mixing and the L-BFGS method for QIS
  and GD, respectively.
  These quasi-Newton techniques exhibit remarkable efficiency gains, resulting
  in orders of magnitude improvements in performance.
  As an application, our algorithms provide a viable approach to designing
  Hamiltonian learning algorithms.
\end{abstract}

\section{Introduction}\label{sec:intro}

% Max entropy

Maximum entropy inference is a widely used method in machine learning,
particularly in the context of graphical
models~\citep{MFP00,KS80,AHS85,Bre15,HKM17} and natural language
processing~\citep{BD-PD-P96}.
In graphical models, it is known as the \emph{backward mapping}, the problem of
computing the model parameters from the marginal information~\citep{WJ07}.
The inverse problem of estimating marginal parameters from the model parameters
is called the \emph{forward mapping}.

% Classical Model Learning

Maximum entropy inference is also a core concept in statistical
physics~\citep{Jay57} known as the Jaynes' principle which links statistical
mechanics and information theory.
The Hammersley-Clifford theorem establishes that, in the classical case, any
positive probability distribution satisfying the local Markov property can be
represented as a Gibbs distribution~\citep{LMP01}.
Hence, maximum entropy inference is equivalent to learning the classical
Hamiltonian defining the Gibbs distribution~\citep{LK20}.
In recent years, sample-efficient and time-efficient algorithms with
demonstrated effectiveness for specific classical graphical models have
emerged~\citep{RWL10,Bre15,HKM17,KM17,VMLC16}.
However, it is technically non-trivial to extend these classical techniques into
the quantum
domain~\citep{WGFC14a,WGFC14b,EHF19,WPS+17,AAKS21,AAR+18,BAL19,BGP+20,HKT22,BLMT23}
due to the non-commutativity of quantum Hamiltonian terms.

% Quantum Hamiltonian Learning

In quantum physics, the maximum entropy inference problem naturally arise when
we only have partial local information about a quantum many-body system.
Let $H_{j}$ for $j=1, 2, \ldots, m$ be local observables of, say, an $n$-qubit
quantum system.
What the state of the system would likely be given only the local marginal
information $\alpha_{j} = \tr(H_{j} \rho)$?
The Jaynes' principle suggests that we take the state to be the one with the
maximum von Neumann entropy, which is always a Gibbs state of the form
$\xi_{H(\mu)} = \frac{1}{Z(\mu)} e^{-\beta H(\mu)}$ where
$H(\mu) = \sum_{j=1}^{m} \mu_{j} H_{j}$ is a quantum Hamiltonian parameterized
by a vector of real numbers $\mu = {(\mu_{j})}_{j=1}^{m}$.
That is, given as input the average values $\alpha = {(\alpha_{j})}_{j=1}^{m}$
encoding local information, the parameter vector $\mu$ of the Hamiltonian is
uniquely determined by $\alpha$ and the computation of $\alpha$ from $\mu$ is an
example of the backward mapping problems.

\begin{figure}[bht]
  \centering
  \begin{subfigure}{}
    \ul{Maximum entropy problem}
    \begin{equation*}
      \begin{split}
        \text{maximize:}\quad & S(\rho)\\
        \text{subject to:}\quad & \ip{H_j}{\rho} = \alpha_{j},\\
                              & \rho \text{\ is a density matrix},
      \end{split}
    \end{equation*}
  \end{subfigure}
  \begin{subfigure}{}
    \ul{Dual problem}
    \begin{equation*}
      \begin{split}
        \text{minimize:}\quad & \ln\tr \exp \Bigl( \sum_{j} \lambda_{j} H_{j} \Bigr) -
                                \lambda \cdot \alpha\\
        \text{subject to:}\quad & \lambda_{j} \in \real.
      \end{split}
    \end{equation*}
  \end{subfigure}
  \caption{The maximum entropy problem and its dual problem.
    Here, $S(\rho) = -\tr(\rho \ln \rho)$ is the von Neumann entropy of
    $\rho$.}\label{fig:max-entropy-and-dual}
\end{figure}

Algorithms for solving the maximum entropy inference problem include the
generalized iterative scaling method~\citep{DR72,D-PD-PL97} which has a rich
history of study in statistics and gradient descent and variable metric methods
originated from optimization research~\citep{Mal02}.
In this paper, we study quantum generalizations of both the Generalized
Iterative Scaling (GIS) algorithm and quasi-Newton methods for solving the
Hamiltonian inference problem.
Overall, the paper offers three key contributions.
Firstly, we present a convergence rate analysis for both Quantum Iterative
Scaling (QIS) and Gradient Descent (GD) algorithms.
This is achieved by representing the iteration's Jacobian in a concise and
explicit formula, which extends a fundamental result from \citet{LYT04}.
Secondly, we establish bounds on the eigenvalues of the Jacobian, proving
polynomial convergence of both the QIS and GD algorithms.
This contribution represents the main technical novelty of the paper.
The non-commutativity of the Hamiltonian terms makes the convergence analysis
highly non-trivial and different from classical proof techniques.
We bound the Hessian of the log-partition function using the monotonicity of the
Bogoliubov--Kubo--Mori (BKM) canonical-correlation form and a lower bound
from~\citet{AAKS21}.
These bounds control the spectrum of the QIS Jacobian even though the Hessian
does not have the simple form available in the classical setting.
Lastly, we investigate two variants of quasi-Newton methods that accelerate the
iterative processes for QIS and GD, respectively, resulting in significant
performance improvements.
In the following, we provide a detailed discussion of these three contributions,
highlighting their technical significance and implications.

\paragraph{Explicit Jacobian Formula for Iterations.}

The QIS algorithm we consider here is a natural quantum counterpart to the GIS
algorithm.
When the matrices are all diagonal, the QIS algorithm reduces to the GIS
algorithm for classical maximum entropy inference.
Previous work~\citep{Ji22} has shown that the algorithm indeed converges to the
correct solution of the maximum entropy problem using the auxiliary function
method and matrix inequalities.
However, while the auxiliary function method is versatile, it falls short in
providing a precise assessment of the algorithms' convergence rates.
To solve this problem, we perform a more tailored analysis of the QIS
algorithm's convergence rate in this paper.
In the classical case, a corresponding convergence analysis for the GIS
algorithm is detailed in~\citet{LYT04}.
This classical analysis leverages Ostrowski's theorem, which bounds the
convergence rate of an iterative procedure by the spectral radius of the
Jacobian matrix associated with the iteration.
Our contribution extends the classical analysis to the quantum setting and
addresses the difficulty that the gradient of the exponential function for
matrices is much more involved than the classical counterpart.
We provide a closed-form formula for the Jacobian of the QIS iteration,
generalizing the approach of~\citet{LYT04}.
The formula has a concise form $\I - P^{-1}L$ where $P$ is a diagonal matrix
whose entries are the mean values of operators related to $H_{j}$ over the Gibbs
state and $L$ is the Hessian of the log-partition function.
This connection to the Hessian matrix $L$ is crucial for the convergence proof.

% GD

The maximum entropy problem has a dual problem, which is an \emph{unconstrained}
optimization problem regarding the log-partition function
in~\cref{fig:max-entropy-and-dual}.
As a comparison, we consider the gradient descent (GD) algorithm for the dual
problem.
The Jacobian of the GD update process can be computed as $\I - \eta L$ where $L$
is the Hessian of the log-partition function and $\eta$ is the step size of
GD\@.
In a sense, the Jacobian of the QIS update rule, $\I - P^{-1} L$, can be seen as
a mechanism to \emph{adaptively} choose the step size for different directions.
This is the main advantage that QIS has over the GD algorithm.
Numerical simulations in \cref{fig:qis-gd} of \cref{sec:experiments} also show
that QIS converges significantly faster than GD\@.

\paragraph{Upper and Lower Bounds for the Jacobian.}

As our main technical contribution, we analyze the eigenvalues of the
Jacobian matrix by establishing both the lower and upper bounds for them.

First, we prove that all eigenvalues of the Jacobian are non-negative by showing
the \emph{upper bound} $L \preceq P$ for the Hessian of the log-partition
function.
The bound follows from monotonicity of the BKM canonical-correlation form under
a completed quantum-to-classical measurement channel, as proved
in~\cref{sec:bounds} using the result of~\citet{Pet94}.

Second, we show that the largest eigenvalue of the Jacobian is bounded away from
$1$ using the \emph{lower bound} on the Hessian from~\citet{AAKS21}, recalled
in~\cref{sec:lower-bound}.
The upper and lower bounds of the Jacobian together complete the convergence
rate analysis of the QIS algorithm by using Ostrowski's theorem
(\cref{thm:ostrowski}).
As a corollary, this shows that for local Hamiltonians satisfying certain
technical conditions, the backward mapping problem can be reduced to the forward
mapping problem efficiently.
The forward mapping problem is unfortunately a hard problem in general and one
may have to resort to approximate inference methods such as variational
inference~\citep{WJ07,CGP19} or Markov entropy decomposition~\citep{GJ07,PH11}

\paragraph{Accelerations.}

% Our results: Anderson

While the QIS and GD algorithms enjoy provable convergence analysis and are
expected to converge in a polynomial number of iterations for local Hamiltonians
at constant temperature, it can exhibit sluggish performance in practical
scenarios.
For classical learning of model parameters, quasi-Newton methods are recommended
for solving the maximum entropy inference problems, as suggested in a systematic
comparison of classical maximum entropy inference algorithms performed
in~\citet{Mal02}.
Even though the convergence analysis is usually less established, quasi-Newton
methods are usually much faster in practice than iterative scaling and gradient
descent algorithms.

In light of this, we investigate two families of quasi-Newton accelerations of
the algorithms.
The first family of heuristic acceleration is based on the Anderson mixing
method~\citep{And65}.
The Anderson mixing algorithm is a heuristic method for accelerating slow
fixed-point iterative algorithms.
It can be seamlessly integrated to work with the QIS algorithm as QIS is indeed
a fixed-point iteration.
The Anderson mixing accelerated QIS algorithm (AM-QIS) has \emph{exactly the
  same} computational requirement of the QIS algorithm in terms of the oracle
access and the type of measurements required on the quantum system.
The second family is based on the BFGS method~\citep{NW06,Yua15} and in
particular the limited memory variant, L-BFGS is applied to the GD algorithm
(L-BFGS-GD).
In our numerical simulations, we observed that AM-QIS and L-BFGS-GD have
comparable performance, usually faster by orders of magnitude than the standard
QIS and GD algorithms.

We believe that applying such quasi-Newton heuristics is important for quantum
optimization algorithms like the QIS algorithm considered here.
While quantum computing offers a promising new paradigm with the potential for
substantial speedups in specific problems, the practical construction of
large-scale quantum computers is still in its early stages.
Current quantum computing technology has limitations in terms of scale and
suffers from errors.
Hence, quantum computing power remains a scarce and valuable resource.
Given this situation, the careful optimization of resources required to solve
problems on quantum computers emerges as a critical task.
The use of Anderson mixing and BFGS for Hamiltonian inference algorithms and
potentially for other fixed-point iterative quantum algorithms represents an
attempt to achieve such resource optimization.
Notably, this approach is not unique to quantum computing.
In fact, the quasi-Newton method, which developed into an important optimization
heuristics, was initially developed by W.\ Davidon while working with early
classical computers, which often crashed before producing correct results.
In response, he devised faster heuristics to expedite calculations later known
as the first quasi-Newton method!
Quantum computers are currently in its very early stages.
They are unstable and prone to errors just like classical computers in the early
days; hence, such heuristic speedups may be critical for numerical quantum
algorithms.

\section{Preliminary}\label{sec:pre}

In this section, we introduce some notations used in this paper.
For two real vectors $x, y \in \real^{m}$, we define $x \cdot y$ as
$\sum_{i=1}^{m} x_{i} y_{i}$.
We sometimes extend this notation to the case when $y$ is a vector of matrices
and write, for example, $\lambda \cdot F$ to mean the summation
$\sum_{j} \lambda_{j} F_{j}$.
For matrices $A, B$, define $\ip{A}{B} = \tr(A^{\dagger} B)$.
We use $A \succeq B$ or $B \preceq A$ to mean $A - B$ is a positive semidefinite
matrix.
A density matrix $\rho$ is a positive semidefinite matrix of unit trace.
The set of density matrices on Hilbert space $\X$ is denoted $\Density(\X)$.

Suppose $\X$ is a finite-dimensional Hilbert space and $f$ is a real convex
function.
We use $\Delta$ to denote the domain $\dom{f}$ of $f$, the interval on which $f$
takes well-defined finite values.
Then $f$ extends to all Hermitian operators in $\Herm_{\Delta}(\X)$ as
$f(X) = \sum_{k} f(\lambda_{k}) \Pi_{k}$ where
$X = \sum_{k} \lambda_{k} \Pi_{k}$ is the spectral decomposition of $X$.
Denote the interior and boundary of $\Delta$ as $\Delta_{\lint}$ and
$\Delta_{\lbd} = \Delta \setminus \Delta_{\lint}$ respectively.
It is easy to see that the domain of matrix function $f$ is
$\Herm_{\Delta}(\X)$, and the interior of the domain is
$\Herm_{\Delta_{\lint}}(\X)$.
For a subset $S$ of Hermitian matrices, we use $\cl{S}$ to denote the closure of
$S$.

Given convex function $f$ as above, the Bregman divergence for matrices is
$D_{f}(X, Y) = \tr(f(X) - f(Y) - f'(Y)(X-Y))$, where $X\in \Herm_{\Delta}(\X)$
and $Y \in \Herm_{\Delta_{\lint}}(\X)$.
An important case we focus on in this paper is $f(x) = x\ln x - x$.
In this case, the matrix Bregman divergence becomes the Kullback-Leibler
divergence $D(X, Y) = \tr(X\ln X - X \ln Y - X + Y)$ defined for non-normalized
matrices $X, Y$.
When $X, Y$ are positive semidefinite matrices of trace $1$, it recovers the
Kullback-Leibler relative entropy $D(X, Y) = \tr(X \ln X - X \ln Y)$.
We will need the matrix Bregman-Legendre projection $\L(Y, \Lambda)$ and
Bregman-Legendre conjugate $\ell(Y, \Lambda)$ for convex function
$f(x) = x \ln x - x$ defined as $\L(Y, \Lambda) = \exp(\ln Y + \Lambda)$,
$\ell(Y, \Lambda) = \tr\exp(\ln Y + \Lambda) - \tr Y$.
For $Y \propto \I$, $\ell(Y, \Lambda) = \tr \exp(\Lambda)$ and we omit $Y$ and
write it as $\ell(\Lambda)$.

In this paper, we consider spin Hamiltonians only and write them as
$H = \sum_{j=1}^{m} H_{j}$ where $H_{j}$'s are local terms acting on at most
constant number of neighboring qubits according to certain interaction geometry.
For example, a $\sigma_{z} \otimes \sigma_{z}$ term acting on the first two
qubits is $\sigma_{z} \otimes \sigma_{z} \otimes \bigl( \I^{\otimes n-2} \bigr)$
for Pauli operator $\sigma_{z} =
\begin{pmatrix}
  1 & 0\\ 0 & -1
\end{pmatrix}
$.
We often use $\xi_{H} = \frac{1}{Z} e^{-\beta H}$ to represent the Gibbs state
of the Hamiltonian $H$ for inverse temperature $\beta$ specified in the context
or $\beta = 1$ otherwise.
Here $Z = \tr e^{-\beta H}$ is the partition function normalizing the state to
have trace $1$ and plays an important role in statistical physics and also in
our work.

\section{Quantum Iterative Scaling}\label{sec:qis}

This section presents a version of the Quantum Iterative Scaling (QIS) algorithm
and discusses its applications in the Hamiltonian inference problem.

We first introduce some notations used in the following discussions.
For a given list of Hermitian matrices $F = {(F_{j})}_{j=1}^{m}$, define the
linear family of quantum states $\lf(\rho_{0})$ as
$\bigl\{ \rho \succeq 0 \mid \ip{F_{j}}{\rho} = \ip{F_{j}}{\rho_{0}} \bigr\}$.
Define the exponential family $\ef(\sigma_{0})$ as
$\bigl\{ \frac{1}{Z} \exp(\ln \sigma_{0} + \lambda \cdot F) \bigr\}$.
We introduce the new notation $F_{j}$ playing the role of $H_{j}$ in the previous
discussion as we will need certain normalization conditions.
In the end, we will choose $F_{j} = \frac{H_{j} + \I}{2m}$ so $F_{j}$ is a
scaled linear shift of $H_{j}$ such that $F_{j} \succeq 0$ and
$\sum_{j} F_{j} \preceq \I$.

We note that, in~\citet{Ji22}, the algorithms are designed for non-normalized
matrices and, therefore, there is no need to explicitly normalize $Y^{(t)}$ in
the update.
Here, we perform explicit normalization to work with normalized quantum states
and their von Neumann entropy.
For $Y^{(t)} = \exp(\ln \sigma_{0} + \lambda \cdot F)$, the normalization is
equivalent to a linear update in the summation of the exponential function
$Y^{(t)} = \exp(\ln \sigma_{0} + \lambda \cdot F - \ln Z)$ where
$Z = \tr Y^{(t)}$.
Hence, we have the following two methods to handle the normalization.
The first is to let the algorithm to find the normalization implicitly, and this
would require that $\I$ is in the span of the $F_{j}$'s.
This will be the case if the assumption on $F_{j}$ is that
$\sum_{j} F_{j} = \I$.
The second is to perform the normalization explicitly as we did in the
algorithm.
This approach is advantageous as it works for all $F_{j}$'s satisfying
$\sum_{j} F_{j} \preceq \I$ even if $\I$ is not in the span of $F_{j}$'s.

An important special case of the algorithm is when $\sigma_{0} = \I/d$ and
$D(\rho, \sigma_{0}) = \ln(d) - S(\rho)$ where $d$ is the dimension.
Then, the minimization over the linear family is now exactly the maximum entropy
problem as in \cref{fig:max-entropy-and-dual} with $H_{j} = F_{j}$,
$\alpha_{j} = \ip{F_{j}}{\rho_{0}}$, and $\lambda_{j} = - \beta \mu_{j}$.
When all the operators $F_{j}$'s are diagonal in the computational basis, the
QIS algorithm recovers the GIS algorithm (see e.g.\ Theorem 5.2
of~\citet{CS04}).

\begin{algorithm}[hbt!]
  \textbf{Require:} $\rho_{0}, \sigma_{0} \in \Density(\X)$ such that
  $D(\rho_{0}, \sigma_{0}) < \infty$.\\
  \textbf{Input:} $F=(F_{1}, F_{2}, \ldots, F_{k}) \in {\Pos(\X)}^{k}$
  and $\sum_{j=1}^{k} F_{j} \preceq \I$.\\
  \textbf{Output:} $\lambda^{(1)}, \lambda^{(2)}, \cdots$ such that
  \begin{equation*}
    \lim_{t\to\infty} D \bigl( \rho_{0}, \L(\sigma_{0},
    \lambda^{(t)} \cdot F) \bigr) = \inf_{\lambda \in \real^{k}} D
    \bigl( \rho_{0}, \L(\sigma_{0}, \lambda \cdot F) \bigr).
  \end{equation*}
  \begin{algorithmic}[1]
    \STATE{Initialize $\lambda^{(1)} = (0, 0, \ldots, 0)$. } %
    \FOR{$t = 1, 2, \ldots, $} %
      \STATE{Compute $Y^{(t)} = \L(\sigma_{0}, \lambda^{(t)} \cdot F)$.}
      \FOR{$j = 1, 2, \ldots, k$} %
        \STATE{$\delta^{(t)}_{j} = \ln \ip{F_{j}}{\rho_{0}} -
          \ln \ip{F_{j}}{Y^{(t)} / \tr Y^{(t)}}$.}
      \ENDFOR{} %
      \STATE{Update parameters
        $\lambda^{(t+1)} = \lambda^{(t)} + \delta^{(t)}$.
      } %
    \ENDFOR{}
  \end{algorithmic}
  \caption{Quantum iterative scaling algorithm.}%%
  \label{alg:qis}
\end{algorithm}

To give some intuition behind the QIS algorithm, we define
$\xi^{(t)} = Y^{(t)} / \tr Y^{(t)}$ and note that the update in the QIS
algorithm is simply
$\delta_{j} = \ln \ip{F_{j}}{\rho_{0}} - \ln \ip{F_{j}}{\xi^{(t)}}$, which is
zero when the linear family constraint $\ip{F_{j}}{\rho} = \ip{F_{j}}{\rho_{0}}$
is satisfied by $\rho = \xi^{(t)}$.
In this case, the algorithm stops updating $\lambda$ in the $j$-th direction as
expected.
Otherwise, if the difference between the current mean value
$\ip{F_{j}}{\xi^{(t)}}$ and the target value $\ip{F_{j}}{\rho_{0}}$ is big, so
will be the update $\delta_{j}$.
The algorithm is, in this sense, \emph{adaptive} when compared to algorithms like
multiplicative weight update algorithms~\citep{AHK12}.

% dual

The maximum entropy problem has dual program in \cref{fig:max-entropy-and-dual}
which is an unconstrained problem.
Hence, it is also attractive to work with the dual using the gradient descent
method (or corresponding quasi-Newton methods discussed later in the paper).
The gradient of the dual objective function is
$\pdv*{\bigl(\ln\ell(\lambda \cdot F) - \lambda \cdot \alpha \bigr)}{\lambda_{j}}
= \ip{F_{j}}{\xi_{\lambda \cdot F}} - \alpha_{j}$,
where $\xi_{\lambda \cdot F}$ is the Gibbs state for Hamiltonian
$\lambda \cdot F$.
Therefore, in gradient descent, the update in each step is
$\eta \bigl( \alpha_{j} - \ip{F_{j}}{\xi_{\lambda \cdot F}} \bigr)$, where
$\eta$ is the learning rate.
This leads to the gradient descent algorithm in \cref{alg:gd}.

\begin{algorithm}[hbt!]
  \textbf{Require:} $\rho_{0}, \sigma_{0} \in \Density(\X)$ such that
  $D(\rho_{0}, \sigma_{0}) < \infty$.\\
  \textbf{Input:} $F=(F_{1}, F_{2}, \ldots, F_{k}) \in {\Pos(\X)}^{k}$
  and $\sum_{j=1}^{k} F_{j} \preceq \I$.\\
  \textbf{Output:} $\lambda^{(1)}, \lambda^{(2)}, \cdots$ such that
  \begin{equation*}
    \lim_{t\to\infty} D \bigl( \rho_{0}, \L(\sigma_{0},
    \lambda^{(t)} \cdot F) \bigr) = \inf_{\lambda \in \real^{k}} D
    \bigl( \rho_{0}, \L(\sigma_{0}, \lambda \cdot F) \bigr).
  \end{equation*}
  \begin{algorithmic}[1]
    \STATE{Initialize $\lambda^{(1)} = (0, 0, \ldots, 0)$.} %
    \FOR{$t = 1, 2, \ldots, $} %
      \STATE{Compute $Y^{(t)} = \L(\sigma_{0}, \lambda^{(t)} \cdot F)$.}
      \FOR{$j = 1, 2, \ldots, k$} %
        \STATE{$\delta^{(t)}_{j} = \eta \ip{F_{j}}{\rho_{0}} -
          \eta \ip{F_{j}}{Y^{(t)}/\tr Y^{(t)}}$.}
      \ENDFOR{} %
      \STATE{Update parameters
        $\lambda^{(t+1)} = \lambda^{(t)} + \delta^{(t)}$.
      } %
    \ENDFOR{}
  \end{algorithmic}
  \caption{Gradient descent algorithm for Kullback-Leibler divergence
    minimization.}%%
  \label{alg:gd}
\end{algorithm}

The QIS algorithm generally outperforms the bare-bones GD algorithm and avoids
the problem of choosing the step size completely.
Consider the update of the QIS algorithm
$\ln \alpha_{j} - \ln \ip{F_{j}}{\xi_{\lambda \cdot F}}$, and the update of the
GD algorithm $\eta \bigl(\alpha_{j} - \ip{F_{j}}{\xi_{\lambda \cdot F}}\bigr)$,
For $\alpha_{j}$ and $\ip{F_{j}}{\xi_{\lambda \cdot F}}$ in
$\interval[open left]{0}{1}$, the QIS update is more aggressive than the dual
gradient descent for learning rate $\eta\le 1$ while still guarantees the
convergence.
This effect is more evident when the two numbers $\alpha_{j}$ and
$\ip{F_{j}}{\xi_{\lambda \cdot F}}$ are small, which holds in most applications.
We will later see that choosing an appropriate learning rate will improve the
performance of the GD algorithm considered in \cref{sec:convergence}, but it is
still less efficient compared to QIS\@.

\section{Convergence Rate}\label{sec:convergence}

In this section, we analyze the geometric convergence rate for the QIS algorithm
in \cref{alg:qis} for the case when $\sigma_{0} = \I/d$.
As a comparison, we will also analyze the convergence rate of the GD algorithm
(\cref{alg:gd}).

We will come across several matrices which are defined here for later references.
For $\lambda$ and $F$, we define the corresponding Gibbs state as
$\xi = \dfrac{\exp(\lambda \cdot F)}{\tr \exp(\lambda \cdot F)}$, and for an
operator $O$, we use $\val{O} = \tr(O \xi)$ to mean the average value of $O$
with respect to $\xi$.
Define diagonal matrix $P = \sum_{j} \val{F_{j}} \ket{j}\bra{j}$.
Finally, define $L$ to be the Hessian of the log-partition function
$\ln \tr \exp(\lambda \cdot F)$ with $\lambda$ as the variables.

We have the following two results regarding the QIS and GD algorithms
respectively, proved in \cref{sec:proofs}.
\begin{theorem}\label{thm:jacobian-qis}
  The Jacobian of the iterative update map
  $\lambda^{(t)} \mapsto \lambda^{(t+1)}$ of \cref{alg:qis} for
  $\sigma_{0} = \I/d$ is given by $\I - P^{-1} L$ for $P$ and $L$ defined above
  with $\lambda = \lambda^{(t)}$.
\end{theorem}

\begin{theorem}\label{thm:jacobian-gd}
  The Jacobian of the iterative update map
  $\lambda^{(t)} \mapsto \lambda^{(t+1)}$ of \cref{alg:gd} is given by
  $\I - \eta L$ where $L$ is the Hessian of the log-partition function defined
  above.
\end{theorem}

By the Ostrowski theorem stated in \cref{thm:ostrowski}, the geometric
convergence rate of the QIS algorithm is, therefore, governed by the spectral
radius of the Jacobian $\I - P^{-1} L$.
Hence, we need to prove bounds on the spectral radius.
In~\citet{AAKS21}, a non-trivial lower bound on $L$ is proved (see
\cref{thm:strong-convexity}), giving an upper bound on the spectral radius.
The other side of the spectral estimate uses the following result, proved in
Appendix~\ref{sec:bounds}.

\begin{theorem}\label{thm:upper-bound}
  Let $P$ and $L$ be matrices defined above. We have $L \preceq P$.
\end{theorem}

We are now able to state the convergence rate of the QIS and GD algorithms.

\begin{theorem}\label{thm:qis-convergence}
  For Hamiltonian $H = \sum_{j = 1}^{m} \mu_{j} H_{j}$ where conditions of
  \cref{thm:strong-convexity} are satisfied and $H_{j}$ are traceless terms with
  norm $\norm{H_{j}} \le 1$, the QIS algorithm with
  $F_{j} = \frac{\I + H_{j}}{2m}$ and the GD algorithm with the same choices of
  $F_{j}$ and $\eta =m$ solve the Hamiltonian inference problem with geometric
  convergence rate $1 - \Omega \bigl( \frac{1}{m^{2}} \bigr)$.
\end{theorem}

\begin{proof}[Proof of \cref{thm:qis-convergence}]
  We consider the QIS algorithm first, which is the more difficult case.
  By \cref{thm:jacobian-qis}, $J_{\QIS} = \I - P^{-1}L$.
  Hence, we can bound the spectral radius as
  \begin{equation*}
    \begin{split}
      r \bigl( J_{\QIS} \bigr)
      & = r\bigl( \I - P^{-1}L \bigr)\\
      & = r\bigl( \I - L^{1/2}P^{-1}L^{1/2}\bigr)\\
      & = \norm{ \I - L^{1/2}P^{-1}L^{1/2} }\\
    \end{split}
  \end{equation*}
  Here, the second line follows from the fact that $AB$ and $BA$ have the same
  set of eigenvalues and the third step follows as $L^{1/2} P^{-1} L^{1/2}$ is
  Hermitian.
  By definition, $P$ is a diagonal matrix whose $(j,j)$-th entry is
  $\ip{F_{j}}{\xi} = \ip{\frac{\I + H_{j}}{2m}}{\xi} \le 1/m$.
  \Cref{thm:upper-bound} then implies that $\I - P^{-1}L$ has eigenvalues in
  $\interval{0}{1}$ and
  \begin{equation}
    \label{eq:qis-jacobian-bound}
    r \bigl( J_{\QIS} \bigr) = 1 - \lambda_{\min}(L^{1/2} P^{-1} L^{1/2})
    \le 1 - m \lambda_{\min} (L).
 \end{equation}

  The Hamiltonian is $H(\mu) = \sum_{j} \mu_{j} (2mF_{j} - \I)$.
  Define $\lambda_{j} = 2m\mu_{j}$, and
  $\tilde{H}(\lambda) = \sum_{j} \lambda_{j} F_{j}$.
  We have $\tilde{H}(\lambda) = H(\mu) + \mu_{\Sigma} \I$ where
  $\mu_{\Sigma} = \sum_{j} \mu_{j} = \sum_{j}\lambda_{j} / (2m)$.
  We compute
  \begin{equation*}
    \begin{split}
      L_{j, k} & = \pdv{\ln\tr \exp (\lambda \cdot F)}
                 {\lambda_{j}, \lambda_{k}}\\
               & = \pdv{\ln\tr \exp (\lambda \cdot F)}{\mu_{j},
                 \mu_{k}} \pdv{\mu_{j}}{\lambda_{j}} \pdv{\mu_{k}}{\lambda_{k}}\\
               & = \frac{1}{4m^{2}}\pdv{ \bigl( \ln\tr\exp (H(\mu)) +
                 \sum_{j} \mu_{j} \bigr)}{\mu_{j}, \mu_{k}}\\
               & = \frac{1}{4m^{2}} \nabla^{2}_{\mu} \ln\tr\exp(H(\mu)).
    \end{split}
  \end{equation*}
  By \cref{thm:strong-convexity}, we have
  $\lambda_{\min}(L) \ge \Omega \Bigl(\frac{1}{m^{3}} \Bigr)$.
  Together with \cref{eq:qis-jacobian-bound}, this completes the proof using
  \cref{thm:ostrowski}.

  By a similar calculation, we can prove the claim for the GD algorithm.
\end{proof}

The above analysis shows that the QIS algorithm has a better geometric
convergence rate even if we set $\eta = m$ in the GD algorithm.
Numerical simulations in \cref{sec:experiments} also confirm this observation.
In some sense, the QIS algorithm is an adaptive gradient descent that can
automatically choose the appropriate learning rate for different dimensions as
$\val{F_{j}}$ may differ for each $j$.

\section{Acceleration by Quasi-Newton Methods}\label{sec:acceleration}

% Intro

The convergence analysis in \cref{sec:convergence} is of theoretical interest
but polynomial convergence proved there is usually not enough for practical
applications.
In this section, we explore the application of quasi-Newton methods, which can
significantly improve the efficiency of the adaptive learning algorithms
considered in \cref{sec:qis}.
In particular, we study two families of methods, the Anderson
mixing~\citep{And65} method and the BFGS method~\citep{NW06}.

% Anderson mixing

Anderson mixing (abbreviated as AM in the following) is a widely used method
employed in numerical and computational mathematics to accelerate the
convergence of fixed-point iterations.
It particularly excels in scenarios where traditional iterative methods may
converge slowly or struggle to find solutions efficiently.
The essence of the Anderson mixing algorithm lies in its ability to dynamically
combine and update a finite set of historical iterates.
It adaptively selects a linear combination of these historical
iterates, leveraging the past information to guide the algorithm toward
convergence more effectively.
This technique finds applications in various scientific and engineering domains,
including quantum chemistry~\citep{GS12}, machine
learning~\citep{SWL+21}, and solving complex systems of
equations~\citep{BCRS22}, where it often delivers substantial
acceleration in computational tasks.

% Use in QIS

Applying the Anderson mixing method to the Hamiltonian inference problem,
specifically to the Quantum Iterative Scaling (QIS) algorithm, is
straightforward due to the inherent nature of QIS as a fixed-point iterative
update algorithm.
The Anderson-accelerated QIS (AM-QIS) algorithm combines both the QIS iterative
step and simple classical processing, so it has exactly the same requirement as
the standard QIS algorithm for the oracle access to the Gibbs state or the
average values.
Since the fixed-point map $g(x)$ in QIS iteration is a contraction, we can set
the mixing parameter $\beta_{t} \equiv 1$ defined in \cref{sec:quasi-newton} and
the convergence of AM-QIS follows from the results in~\citet{TK15}.
We also use the Barzilai-Borwein (BB) method~\citep{BB88} for choosing the
mixing parameter which turns out to be effective and provides further
accelerations.

% BFGS-GD

The BFGS method and the limited memory variant L-BFGS are the most influential
among many quasi-Newton methods.
They are the recommended choice for learning graphical models in the classical
machine learning literature~\cite{Mal02}.
The BFGS method works with an unconstrained optimization problem
$\min_{x\in \real^{n}} f(x)$.
The update in the BFGS algorithm has the form
$x_{k+1} = x_{k} - \eta_{k} H_{k} \nabla f(x_{k})$, where $\eta_{k}$ is the step
size which can usually be found by line search, and $H_{k}$ is a matrix that is
updated iteratively during the execution of the algorithm.
We consider both a fixed choice or the BB method for the initial approximation
of the inverse Hessian $H_{0}$ for a fair comparison with AM\@.
The application of BFGS methods to our problem is also straightforward as the
dual problem is an unconstrained optimization problem.

\begin{figure}[htb]
  \centering
  \includegraphics[width=.9\linewidth]{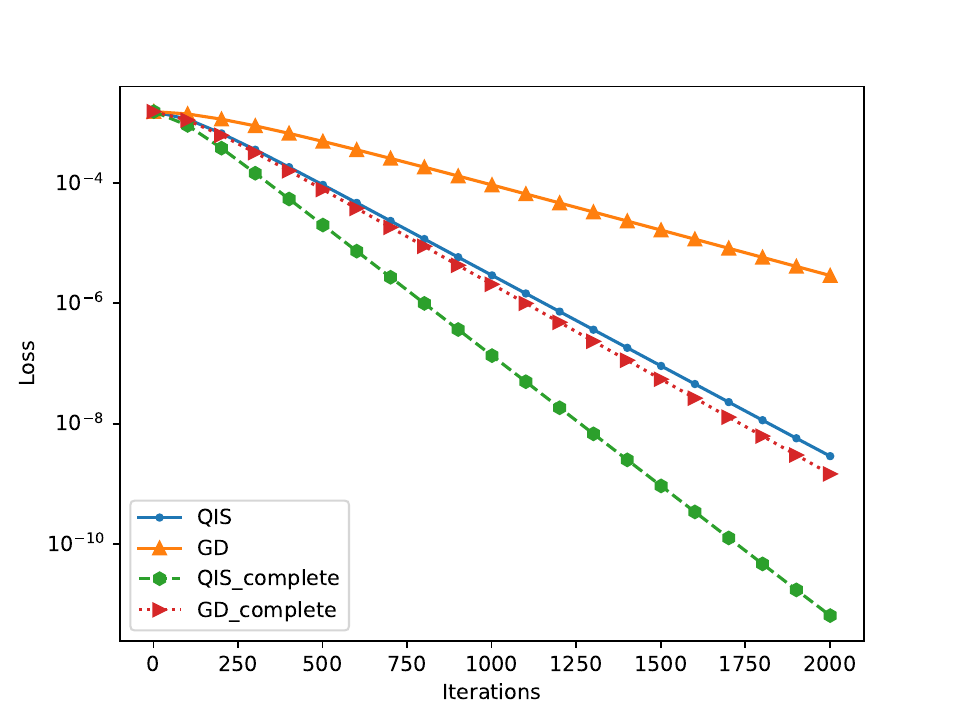}
  \caption{Comparison of QIS and GD algorithms.
    The loss in measured by the error in the objective function of the maximum
    entropy problem.}\label{fig:qis-gd}
\end{figure}

AM and BFGS have different application scenarios.
AM is an acceleration method for solving fixed-point problems and the
approximation $G_t$ of the inverse Jacobian matrix is generally not symmetric.
In contrast, BFGS is an optimization method and constructs a symmetric
approximation $H_t$ for the inverse Hessian matrix.
Since an optimization problem can usually be recast as a fixed-point problem, AM
also applies to solving optimization problems.
However, BFGS may be more efficient in some cases due to the maintained symmetry
structure compared with AM\@.

\section{Experiments}\label{sec:experiments}

We conducted numerical simulations to assess the comparative efficiency of four
approaches: the standard QIS, the standard GD algorithm, AM-QIS, and the
L-BFGS-GD algorithm applied to the dual problem.
The Linux workstation we used for the numerical experiment has a 16-core CPU
(Intel(R) Xeon(R) Platinum 8369HB CPU @ 3.30GHz) and 64GB of memory.

In the experimental setup, we adopted a method involving generating random Gibbs
states for random local Hamiltonians, represented as
$H = \sum_{j} \lambda_{j} H_{j}$.
Here, the local terms, denoted as $H_{j}$, consist of tensor products of local
Pauli operators, and the $\lambda_{j}$ parameters are the values to be learned.
These Hamiltonians were then utilized to create Gibbs states
$\xi_{H} = \frac{1}{Z} e^{-\beta H}$.
We feed the Gibbs states and their corresponding local average values
$\alpha_{j} = \ip{H_{j} \otimes \I}{\xi_{H}}$ to the algorithms.
In this way, we know the ground truth about the values of $\lambda_{j}$'s and
the objective value of the optimization programs in
\cref{fig:max-entropy-and-dual}, and we choose to evaluate the algorithms'
performance by the error compared with the true objective value.

\begin{figure}[ht]
  \centering
  \subfigure[With BB method.]{
    \includegraphics[width=0.45\textwidth]{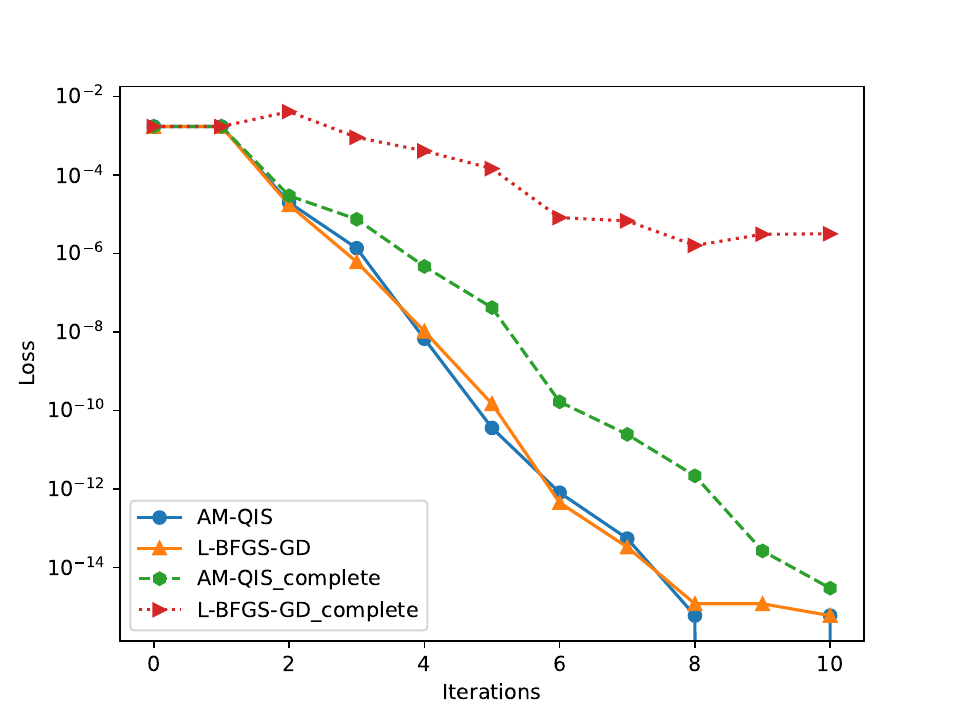}
  }\label{fig:amqis-lbfgs-a} \subfigure[Without BB method.]{
    \includegraphics[width=0.45\textwidth]{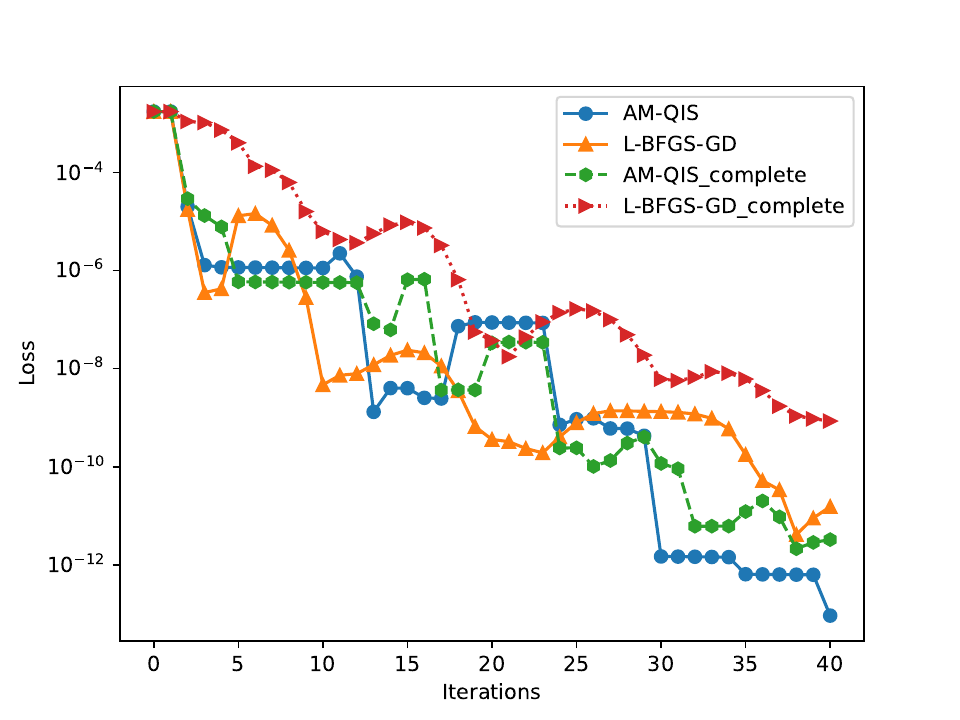}
  }\label{fig:amqis-lbfgs-b}
  \caption{Comparisons of the AM-QIS and L-BFGS-GD algorithms.
    The \cref{fig:amqis-lbfgs-a} on the left uses the Barzilai-Borwein method to
    choose the step size and \cref{fig:amqis-lbfgs-b} on the right uses fixed
    step size.
    The dotted (red) and dashed (green) lines represent the performance of the
    algorithms when the input Hamiltonian terms are complete satisfying
    $\sum_{j} F_{j} = \I$.
  }\label{fig:amqis-lbfgs}
\end{figure}

The results are summarized in \cref{fig:qis-gd,fig:amqis-lbfgs}.
In \cref{fig:qis-gd}, we compare the performance of QIS and GD algorithms.
We can see that QIS algorithm is more efficient than GD algorithm regardless of
whether we ensure the completeness $\sum F_{j} = \I$ or not.
In \cref{fig:amqis-lbfgs}, we compare the performance of AM-QIS and L-BFGS-GD,
both with and without the Barzilai-Borwein method.
We can see that AM-QIS and L-BFGS-GD are comparable in general.
The standard QIS algorithm typically required approximately 1500 iterations to
achieve an error level of $10^{-6} \sim 10^{-8}$ (measured using the objective
function of the maximum entropy problem).
In contrast, the AM-QIS and L-BFGS algorithms achieved the same accuracy with
only about 8 (or 20) iterations with (or without) BB, showcasing a remarkable
speedup of two orders of magnitude.
The efficiency of the AM-QIS algorithm is stable and does not change much when
the Hamiltonian is normalized and completed, while the efficiency of L-BFGS-GD
algorithm (in \cref{fig:amqis-lbfgs-a}) is observed to be sensitive in this
regard.

\section{Discussions}

In this study, we considered adaptive learning algorithms for the Hamiltonian
inference problem.
We examined the convergence of the quantum iterative scaling algorithm (QIS) and
the gradient descent (GD) algorithm for the dual problem.
Furthermore, two quasi-Newton methods AM-QIS and L-BFGS-GD
are proposed.

The QIS algorithm iteratively updates the Hamiltonian parameters adaptively by
comparing $\ip{H_{j} \otimes \I}{\xi_{H(\lambda)}}$ and the target value
$\alpha_{j}$.
Therefore it requires the ability to prepare $\xi_{H(\lambda)}$ for the trial
parameter $\lambda$ or estimate the local information of the state.
This is generally a computationally demanding assumption, but if the physical
system has exponentially decaying correlation and satisfy certain Markov
property, the preparation of the Gibbs state or its local observations could be
efficient~\citep{BK19,KKB20}.
Furthermore, the issue may be solved or mitigated by combining quantum belief
propagation algorithms proposed in~\citet{Has07,LP08,PH11} which present
possible ways of computing the value $\ip{H_{j} \otimes \I}{\xi_{H(\lambda)}}$
approximately without generating the full Gibbs state, thereby removing the use
of the adaptive Gibbs oracle.
We leave the exploration of this possibility as future work.

\newpage
\bibliography{hamiltonian-inference}

\bibliographystyle{natbib-links}

\newpage
\appendix
\onecolumn
\section{Prime and Dual Divergence Minimization Problems}

The primal and dual formulation of the maximum entropy program in
\cref{fig:max-entropy-and-dual} is a special case of the following duality
result between two minimization problems of the Kullback-Leibler divergence
given in \cref{fig:divergence-optimization-problems}.
The duality theorem of~\citet{Ji22}, or Jaynes' principle, states that the
following two problems have the same minimizer which is the unique intersection
point of the linear family $\lf(\rho_{0})$ and exponential family
$\cl{\ef(\sigma_{0})}$ defined in \cref{sec:qis}.
When $\sigma_{0}$ is the maximally mixed state $\I/d$, the linear family
minimization is the maximum entropy problem and the exponential family
minimization is a dual program in \cref{fig:max-entropy-and-dual}.

\begin{figure}[bht]
  \centering
  \begin{subfigure}{}
    \ul{Linear family minimization}
  \begin{align*}
    \text{minimize:}\quad & D(X, \sigma_{0})\\
    \text{subject to:}\quad & X \in \lf(\rho_{0}).
  \end{align*}
  \end{subfigure}
  \begin{subfigure}{}
    \ul{Exponential family minimization}
    \begin{align*}
    \text{minimize:}\quad & D(\rho_{0}, Y)\\
    \text{subject to:}\quad & Y \in \ef(\sigma_{0}).
    \end{align*}
  \end{subfigure}
  \caption{Two optimization problems of the Kullback-Leibler divergence that are
    dual to each other.}\label{fig:divergence-optimization-problems}
\end{figure}

\section{Proofs for Convergence Rate}\label{sec:proofs}

This section proves the explicit formulas for the Jacobian matrix of the
iterations in QIS and GD algorithms.

\begin{proof}[Proof of \cref{thm:jacobian-qis}]
  We first prove two identities about the partial derivative of the function
  $\ell$ and its natural logarithm.
  \begin{align}
    \pdv*{\ell(\lambda \cdot F)}{\lambda_{j'}}
    & = \ip{F_{j'}}{\exp(\lambda \cdot F)}, \label{eq:pdv-ell}\\
    \pdv*{\ln \ell(\lambda \cdot F)}{\lambda_{j'}}
    & = \ip{F_{j'}}{\xi}, \label{eq:pdv-ln-ell}
  \end{align}
  for $\xi$ defined as
  \begin{equation}
    \label{eq:gibbs-xi}
    \xi = \dfrac{\exp(\lambda \cdot F)}{\tr \exp(\lambda \cdot F)}.
  \end{equation}
  In fact, we have
  \begin{equation*}
    \begin{split}
      \pdv*{\ell(\lambda \cdot F)}{\lambda_{j'}}
      & = \pdv*{\tr\, \exp(\lambda \cdot F)}{\lambda_{j'}}\\
      & = \tr\, \sum_{k=0}^{\infty}
        \pdv*{\frac{{(\lambda \cdot F)}^{k}}{k!}}{\lambda_{j'}}\\
      & = \tr\, \sum_{k=1}^{\infty}\sum_{j=0}^{k-1}
        \frac{{(\lambda \cdot F)}^{k-j-1} F_{j'} {(\lambda \cdot F)}^{j}}{k!}\\
      & = \Bigl\langle F_{j'}, \sum_{k=1}^{\infty}
        \frac{{(\lambda \cdot F)}^{k-1}}{(k-1)!} \Bigr\rangle\\
      & = \ip{F_{j'}}{\exp(\lambda \cdot F)},
    \end{split}
  \end{equation*}
  where the second line follows from the Taylor expansion of the matrix
  expansion and the fourth line is by the cyclic property of trace.
  This proves \cref{eq:pdv-ell}.
  Similarly, we have
  \begin{equation*}
    \begin{split}
      \pdv*{\ln \ell(\lambda \cdot F)}{\lambda_{j'}}
      & = \frac{1}{\ell(\lambda \cdot F)} \pdv*{\ell(\lambda \cdot F)}{\lambda_{j'}}\\
      & = \frac{1}{\tr \exp(\lambda \cdot F)} \ip{F_{j'}}{\exp(\lambda \cdot F)}\\
      & = \ip{F_{j'}}{\xi},
    \end{split}
  \end{equation*}
  where the second line follows from \cref{eq:pdv-ell}.
  This completes the proof of \cref{eq:pdv-ln-ell}.

  Now, for $j=1, 2, \ldots, k$, the update in the algorithm is
  \begin{equation*}
    \delta_{j} = \ln \ip{F_{j}}{\rho_{0}} - \ln \ip{F_{j}}{Y^{(t)} / \tr Y^{(t)}}.
  \end{equation*}
  Hence, for $j, j' \in \{1, 2, \ldots, k\}$ and $\xi$ defined in \cref{eq:gibbs-xi},
  \begin{equation*}
    \begin{split}
      \pdv{\delta_{j}}{\lambda_{j'}}
      = & - \frac{1}{\ip{F_{j}}{\xi}}
          \pdv{\ip{F_{j}}{\xi}}{\lambda_{j'}}\\
      = & - \frac{1}{\ip{F_{j}}{\xi}}
          \pdv*{\pdv*{\ln \ell(\lambda \cdot F)}{\lambda_{j}}}{\lambda_{j'}}\\
      = & - P_{j, j}^{-1} L_{j, j'}.
    \end{split}
  \end{equation*}
  For the second line, we used \cref{eq:pdv-ln-ell}.
  Equivalently, the Jacobian matrix
  \begin{equation*}
    {\Bigl(\pdv{\delta_{j}}{\lambda_{j'}} \Bigr)}_{j,j'} = - P^{-1} L,
  \end{equation*}
  for matrices $P$ and $L$ defined in the statement of the theorem.
  The Jacobian $J_{\QIS}$ of the QIS iteration can be written as
  \begin{equation*}
    J_{\QIS} = \I + {\Bigl( \pdv{\delta_{j}}{\lambda_{j'}} \Bigr)}_{j,j'}
    = \I - P^{-1} L.
  \end{equation*}
  This completes the proof the theorem.
\end{proof}

\begin{proof}[Proof of \cref{thm:jacobian-gd}]
  In an iteration of the algorithm, we have
  \begin{equation*}
    \delta_{j} = \eta \ip{F_{j}}{\rho_{0}} - \eta \ip{F_{j}}{Y^{(t)}/\tr Y^{(t)}}.
  \end{equation*}
  The Jacobian $J_{\GD}$ of each iteration has $(j,j')$ entry
  \begin{equation*}
    \begin{split}
      & \I - \eta \, \pdv*{\ip{F_{j}}{\xi}}{\lambda_{j'}} \\
      = \, & \I - \eta \, \pdv*{\pdv*{\ln \tr \exp(\lambda
             \cdot F)}{\lambda_{j}}}{\lambda_{j'}}\\
      = \, & \I - \eta L.
    \end{split}
  \end{equation*}
  This completes the proof.
\end{proof}

We recall a theorem of Ostrowski which we will use to prove the convergence rate
by bounding the spectral radius of a Jacobian matrix.
\begin{theorem}[Ostrowski's theorem~{\citep[Chapter~22]{Ost66}}]\label{thm:ostrowski}
  Assume function $f$ is differentiable at the neighborhood of a fixed point
  $\zeta$.
  For an iterative algorithm $\zeta_{t+1} = f(\zeta_{t})$.
  A sufficient condition for $\zeta$ to be a point of attraction is the
  spectral radius $r(J_f) < 1$.
  Moreover, if $\zeta$ is an attraction point, the geometric convergence rate of
  the iterative algorithm is given by
  \begin{equation*}
    \limsup_{t \to \infty} \frac{\norm{\zeta_{t+1} -
        \zeta}}{\norm{\zeta_{t} - \zeta}} = r(J_{f}).
  \end{equation*}
\end{theorem}

\section{Bounds on the Hessian Matrix}\label{sec:bounds}

\begin{theorem}[Duhamel's formula]%
  \label{thm:duhamel}
  Let $A(s)$ be a matrix-valued function. Then
  \begin{equation*}
   \odv*{\exp(A(s))}{s} = \int_0^1 \odif{t}\; \exp(tA(s)) A'(s) \exp((1-t)A(s)).
  \end{equation*}
\end{theorem}

Define matrices
\begin{align*}
  \Delta
  & = \sum_{j} \pdv{\ell(\lambda \cdot F)}{\lambda_{j}}
      \ket{j}\bra{j},\\
  \Lambda
  & = \sum_{j,j'} \pdv{\ell(\lambda \cdot F)}
      {\lambda_{j}, \lambda_{j'}} \ket{j}\bra{j'}.
\end{align*}
In other words, $\Delta$ is a diagonal matrix and $\Lambda$ is the Hessian of
the partition function
$Z = \ell(\lambda \cdot F) = \tr\exp(\lambda \cdot F)$.
In the following, let
$\rho = \exp(\lambda\cdot F)/\tr\exp(\lambda\cdot F)$.
We use $\langle, \rangle$ for the Frobenius inner product on matrices, i.e.,
$\langle A, B\rangle = \tr(A^{\dagger} B)$, and $\langle, \rangle_{\rho}$
for the BKM inner product
\begin{equation*}
  \langle A,B\rangle_\rho
  = \tr\!\left(
      A^\dagger\int_0^1 \rho^tB\rho^{1-t}\,\odif{t}
    \right).
\end{equation*}

\begin{proposition}\label{prop:entry-gradient-Hessian}
  The $j$-th diagonal entry of $\Delta$ is
  \begin{equation*}
    \Delta_{jj}
    = \langle F_j, \exp(\lambda \cdot F)\rangle
    = Z \langle F_j, I\rangle_{\rho},
  \end{equation*}
  and the $jk$-th entry of $\Lambda$ is
  \begin{equation*}
    \Lambda_{jk}
    = Z \left\langle F_j,
      \int_0^1 \rho^{t} F_k \rho^{1-t}\,\odif{t}\right\rangle
    = Z \langle F_j, F_k \rangle_{\rho}.
  \end{equation*}
\end{proposition}

We recall the Petz monotonicity theorem as follows.

\begin{theorem}[Theorem 2.2 and Equation (4.2) in~\cite{Pet94}]%
  \label{thm:Petz-monotonicity}
  Let $\mathcal{H}_{A}$ and $\mathcal{H}_{B}$ be finite-dimensional Hilbert
  spaces, and let
  $\mathcal{E}\colon \mathcal{L}(\mathcal{H}_{A}) \longrightarrow
  \mathcal{L}(\mathcal{H}_{B})$ be completely positive and trace-preserving.
  Let $\mathcal{E}^{\dagger}\colon \mathcal{L}(\mathcal{H}_{B})
  \longrightarrow \mathcal{L}(\mathcal{H}_{A})$ be its Hilbert--Schmidt
  adjoint.
  For every density operator $\rho \in \Density(\mathcal{H}_{A})$ and every
  $K \in \mathcal{L}(\mathcal{H}_{B})$, it holds that
  \begin{equation*}
    \langle K,K\rangle_{\mathcal{E}(\rho)}
    \ge
    \langle \mathcal{E}^{\dagger}(K),
      \mathcal{E}^{\dagger}(K)\rangle_{\rho}.
  \end{equation*}
  In finite dimensions, all the operators under consideration are bounded.
\end{theorem}

\begin{theorem}\label{thm:dominant}
  For $\Lambda$ and $\Delta$ defined above, we have $\Lambda \preceq \Delta$.
\end{theorem}

\begin{proof}
  Fix an arbitrary $v = (v_1,\ldots,v_n) \in \complex^n$, and define
  \begin{equation*}
    F_0 = \I - \sum_{j=1}^n F_j \succeq 0.
  \end{equation*}
  Complete the measurement by defining the quantum-to-classical channel
  \begin{equation*}
    \widetilde{\Phi}(X)
    = \sum_{j=0}^n \tr(XF_j)\ket{j}\bra{j}.
  \end{equation*}
  The map $\widetilde{\Phi}$ is CPTP because each $F_j \succeq 0$ and
  $\sum_{j=0}^n F_j = \I$.
  Define
  \begin{equation*}
    \widetilde{W}
    = 0\ket{0}\bra{0} + \sum_{j=1}^n v_j\ket{j}\bra{j},
    \qquad
    B = \sum_{j=1}^n v_jF_j,
  \end{equation*}
  and let $\widetilde{p} = \widetilde{\Phi}(\rho)$.
  The Hilbert--Schmidt adjoint satisfies
  \begin{equation*}
    \widetilde{\Phi}^{\dagger}(\widetilde{W})
    = \sum_{j=1}^n v_jF_j
    = B.
  \end{equation*}
  Applying~\Cref{thm:Petz-monotonicity} to the CPTP map
  $\widetilde{\Phi}$ gives
  \begin{equation*}
    \langle \widetilde{W},\widetilde{W}\rangle_{\widetilde{p}}
    \ge \langle B,B\rangle_{\rho}.
  \end{equation*}
  Since the coefficient of outcome $0$ in $\widetilde{W}$ is zero, that
  outcome contributes nothing, and~\Cref{prop:entry-gradient-Hessian} gives
  \begin{align*}
    \langle \widetilde{W},\widetilde{W}\rangle_{\widetilde{p}}
    & = \sum_{j=1}^n \tr(\rho F_j)|v_j|^2\\
    & = \frac{1}{Z}v^{\dagger}\Delta v.
  \end{align*}
  For the other side, the same proposition gives
  \begin{align*}
    \langle B,B\rangle_{\rho}
    & = \left\langle
      \sum_jv_jF_j,\sum_kv_kF_k
      \right\rangle_{\rho}\\
    & = \sum_{j,k}v_j^{*}v_k\langle F_j,F_k\rangle_{\rho}\\
    & = \frac{1}{Z}v^{\dagger}\Lambda v.
  \end{align*}
  Therefore, $v^{\dagger}\Delta v \ge v^{\dagger}\Lambda v$ for every
  $v \in \complex^n$, and hence $\Lambda \preceq \Delta$.
\end{proof}

We now derive the upper Hessian bound in~\cref{thm:upper-bound}.
Let
\begin{equation*}
  Z = \tr\exp(\lambda\cdot F),
  \qquad
  p_j = \langle F_j\rangle_\rho = \tr(\rho F_j).
\end{equation*}
By~\Cref{prop:entry-gradient-Hessian}, $\Delta_{jj} = Zp_j$.
Thus, for the diagonal matrix $P$ defined in~\cref{sec:convergence},
\begin{equation*}
  P = \frac{\Delta}{Z}.
\end{equation*}
For the Hessian $L = \nabla^2_\lambda \ln Z$, we have
\begin{align*}
  L_{jk}
  & = \frac{1}{Z}\pdv{Z}{\lambda_j,\lambda_k}
      - \frac{1}{Z^2}\pdv{Z}{\lambda_j}\pdv{Z}{\lambda_k}\\
  & = \frac{\Lambda_{jk}}{Z} - p_jp_k.
\end{align*}
Thus, in matrix form,
\begin{equation*}
  L = \frac{\Lambda}{Z} - pp^{\mathsf T},
\end{equation*}
where $p = (p_1,\ldots,p_n)^{\mathsf T}$ is real.
Using $\Lambda \preceq \Delta$, we obtain
\begin{equation*}
  L
  \preceq \frac{\Delta}{Z} - pp^{\mathsf T}
  = P - pp^{\mathsf T}
  \preceq P.
\end{equation*}
This proves~\cref{thm:upper-bound}.

\section{Lower Bound on the Hessian Matrix}\label{sec:lower-bound}

For the other side of the spectral estimate, we recall the following lower
bound on the Hessian from~\citet{AAKS21}.
\begin{theorem}[Theorem 6 of~\citet{AAKS21}]\label{thm:strong-convexity}
  Let $H(\mu) = \sum_{j=1}^{m} \mu_{j} H_{j}$ be an $\ell$-local Hamiltonian
  over a finite dimensional lattice.
  For a given inverse temperature $\beta$, there are constants $c, c'> 3$
  depending on the geometric property of the lattice such that
  \begin{equation*}
    \nabla^{2}_{\mu} \ln \tr \bigl( e^{-\beta H(\mu)} \bigr)
    \succeq \frac{e^{-O(\beta^{c})} \beta^{c'}}{m} \I.
  \end{equation*}
\end{theorem}

\section{Discussions on Quasi-Newton Methods}\label{sec:quasi-newton}

In this section, we give some details of the Anderson mixing method and L-BFGS
method.

% Anderson mixing with details

The Anderson mixing method interpolates history information in order to speed up
a fixed-point iteration.
More concretely, suppose $g: \real^{d} \rightarrow \real^{d}$ is a contraction
and we are interested in finding the fix-point $x = g(x)$.
The standard fix-point iterative algorithm is to compute
$x_{t+1} = g \bigl( x_{t} \bigr) \text{ for } t = 0, 1, 2, \ldots$, until a
stopping criteria is met.
In Anderson mixing, a relatively small history size $m \ge 0$ is chosen and we
define $m_{t} = \min\{m, t\}$.
In our numerical implementation, we use $m = 10$.
Define the residual $r_{t} = g \bigl( x_{t} \bigr) - x_{t}$ and two matrices
$X_{t}, R_{t} \in \real^{d\times m}$ storing the historical information
\begin{align*}
  X_{t} & = \bigl( \Delta x_{t-m_{t}}, \Delta x_{t-m_{t}+1}, \ldots, \Delta x_{t-1} \bigr),\\
  R_{t} & = \bigl( \Delta r_{t-m_{t}}, \Delta r_{t-m_{t}+1}, \ldots, \Delta r_{t-1} \bigr),
\end{align*}
where $\Delta$ is the forward difference operator and
$\Delta x_{k} = x_{k+1} - x_{k}$.
Then, the Anderson accelerated iteration can be written succinctly as
$x_{t+1} = x_{t} + G_{t} r_{t}$ where
\begin{equation*}
  G_{t} = \beta_{t} I - (X_{t} + \beta_{t} R_{t})
  {\bigl( R_{t}^{T}R_{t} \bigr)}^{-1} R_{t}^{T}.
\end{equation*}
Here, $\beta_{t}$ is the mixing parameter.

It is pointed out in~\citet{FS09} that $G_{t}$ approximates the inverse of the
Jacobian of $g$ and Anderson mixing method can be thought of as a quasi-Newton
method satisfying multi-secant equations.
We note that there is a matrix inverse in the above formula for $G_{t}$ which
can be implemented using Moore-Penrose pseudo-inverse.
For stability and efficiency concerns, we found that the AM algorithms have the
best performance in our numerical simulations when using a relative condition
number of $1\mathrm{e}{-}7$ in the pseudo-inverse, a cutoff threshold that sets
small singular values of the matrix to zero.
This is easily implemented by setting the $\mathrm{rcond}$ parameter of the
$\mathrm{pinv}$ function in $\mathrm{numpy.linalg}$ package for Python
implementations.

The BFGS method is one of the most popular quasi-Newton methods that can be
applied to unconstrained optimization problems $\min_{x \in \real^{n}} f(x)$.
It is also known as the variable metric algorithm as first proposed by
Davidon~\citep{Dav91,Yua15}.
The algorithm maintains the approximate Hessian $H_{k+1}$ of the optimization
problem.
The update rule of the algorithm is
\begin{align*}
  x_{t+1} = x_t - \eta_t H_t \nabla f(x_t),
\end{align*}
and the update rule of $H_t$ is
\begin{align*}
  H_{t+1} = \Bigl(\I - \frac{s_t y_t^T}{y_t^T s_t} \Bigr) H_t
  \Bigl(\I - \frac{y_t s_t^T}{y_t^T s_t}\Bigr) + \frac{s_t s_t^T}{y_t^T s_t},
\end{align*}
where $s_t = x_{t+1} - x_t$, $y_t = \nabla f(x_{t+1}) - \nabla f(x_t)$ and $H_0$
is a predefined initial approximation of the inverse Hessian matrix.
We refer readers to~\citet{NW06} for a discussion on how the BFGS update is
derived.
To optimize the memory usage, the limited memory version of BFGS called
L-BFGS~\citep{LN89} is used.
Since using line search for choosing $\eta_t$ can incur additional function
evaluations in each iteration, we use $\eta_t \equiv 1$ and only tune $H_0$ in
the numerical simulation.

Both the AM and the BFGS methods employed our numerical simulation can be
further strengthened by using a heuristics invented by Barzilai and
Borwein~\citep{BB88} to choose the $\beta_t$ in AM and $ H_0$ in BFGS\@.
In AM, we set
\begin{align*}
  \beta_{t} = - \frac{\Delta r_{t-1}^T \Delta x_{t-1}}{\Delta r_{t-1}^T \Delta r_{t-1}},
\end{align*}
which solves $\min_{\beta} \norm{\Delta x_{t-1} + \beta \Delta r_{t-1}}_{2}$.
In BFGS, we set
\begin{align*}
  H_{0} = \frac{y_{t-1}^T s_{t-1}}{y_{t-1}^T y_{t-1}} \I
\end{align*}
to be the initialization of the approximate inverse Hessian in the $t$-th
iteration~\citep[Pages 143 and 178]{NW06}.

\section{Experimental Details}\label{sec:experimental-details}
In this section, we provide details of the numerical experiments.

\subsection{Hamiltonian Generation}\label{sub-sec:hamiltonian-generation}

Three types of Hamiltonians are used in our numerical experiments.
These include the Ising family of Hamiltonians, the transversal $2$-local
Hamiltonians in 1D (Transversal1D), and the $2$-local Hamiltonians in 1D
(Local1D).
The figures in the main text demonstrate the results of the experiments on the
Local1D Hamiltonian in $6$-qubit system.

\paragraph{Ising Hamiltonian}
The Ising Hamiltonian is a sum of $2$-local terms, which can be written as
\begin{equation*}
  H = \sum_{i=1}^{n} a_i \sigma_{x}^{(i)} +
  \sum_{i=1}^{n-1} b_i \sigma_{z}^{(i)} \sigma_{z}^{(i+1)},
\end{equation*}
where $\sigma_{z}^{(i)}$ is a Pauli $Z$ operator acting on the $i$-th qubit.
The coefficients $a_i$ and $b_i$ are randomly generated from the standard normal
distribution.

\paragraph{Transversal1D Hamiltonian}
The transversal 1D Hamiltonian is a sum of $2$-local terms, and it can be
written as
\begin{equation*}
  H = \sum_{p\in \{x,y,z\}} a_p \sum_{i=1}^{n} \sigma_{p}^{(i)} + \sum_{p,q \in \{x,y,z\} }
  b_{p,q} \sum_{i=1}^{n} \sigma_{p}^{(i)}\sigma_{q}^{(i+1)}.
\end{equation*}
The coefficients $a_j$ and $b_{p,q}$ are randomly generated from the standard
normal distribution.

Notice that, in this scenario, there are $12$ terms in the Hamiltonian, and each
term in the Hamiltonian is a sum of Pauli operators acting on different qubits.
Thus, the operator norm of each term is bounded by the number of qubits instead
of $1$.
As a result, in the implementation of our algorithm, we need to normalize the
Hamiltonian by dividing the number of qubits.

\paragraph{Local1D Hamiltonian}
The local 1D Hamiltonian is a sum of $2$-local terms, including the Ising
Hamiltonian as a special case, and it can be written as
\[
  H = \sum_{i=1}^{n}\sum_{p\in\{x,y,z\}} a_{p,i} \sigma_{p}^{(i)}
  + \sum_{i=1}^{n}\sum_{p,q\in\{x,y,z\}} b_{p,q,i} \sigma_{p}^{(i)} \sigma_{q}^{(i+1)},
\] 
where $\sigma^{(i)}$ is a Pauli operator acting on the $i$-th qubit for $i\le n$,
and $\sigma^{(n+1)}$ is defined as $\sigma^{(n+1)} = \sigma^{(1)}$.
The coefficients $a_{p,i}$ and $b_{p,q,i}$ are randomly generated from the standard
normal distribution.

\subsection{Initialization, Learning Rate, and Stopping Criteria}\label{sub-sec:learning-rate}
The initial guess for the parameters in the QIS algorithm and the gradient
descent algorithm is set to $0$.
The learning rate in the gradient descent algorithm is set to the number of
terms in the Hamiltonian.
In the experiments, we first generate the Hamiltonian and then normalize it by
dividing it by the number of qubits.
Then, the inverse temperature $\beta$ is set to $1$.
Therefore, the optimal solutions and optimal objective values can be directly
computed and compared with the results of the algorithms.
With these settings, the stopping criteria can be set to the error of the
objective value.
The stopping criteria for the QIS, gradient descent algorithm, and quasi-Newton
accelerations is set to have error at most $10^{-12}$.
The maximum number of iterations for the quasi-Newton accelerations is set to
$40$.

\subsection{Results on Different Hamiltonians}\label{sub-sec:results-hamiltonians}
In this section, we provide the results of the numerical experiments on
different Hamiltonians.
We have conducted experiments on the Ising, Transversal1D, and Local1D
Hamiltonians, with different numbers of qubits.
The results are shown in the following tables and figures.
We first present the results in \cref{tab:experiment-results} and then provide
the figures for each family of Hamiltonians.
\begin{table}[]
  \renewcommand{\arraystretch}{1.5}
  \centering
  \begin{tabular}{lrrcc}
  \toprule
                      & QIS  & GD   & AM-QIS & L-BFGS-GD \\ \midrule
  6-qubit-Ising       & 319  & 479  & 7      & 6         \\
  6-qubit-Transversal & 1743 & 2562 & 5      & 7         \\
  6-qubit-Local       & 1913 & 2377 & 5      & 8         \\
  7-qubit-Ising       & 370  & 551  & 6      & 6         \\
  7-qubit-Transversal & 2006 & 2940 & 5      & 7         \\
  7-qubit-Local       & 2236 & 2721 & 5      & 7         \\
  8-qubit-Ising       & 421  & 623  & 6      & 5         \\
  8-qubit-Transversal & 2267 & 3311 & 4      & 7         \\
  8-qubit-Local       & 2525 & 3020 & 5      & 7         \\\bottomrule
  \end{tabular}
  \caption{Number of steps to achieve $10^{-7}$ precision for different
    Hamiltonians.}\label{tab:experiment-results}
\end{table}

\begin{figure}[ht]
  \centering
  \subfigure[QIS and GD results.]{
    \includegraphics[width=0.6\textwidth]{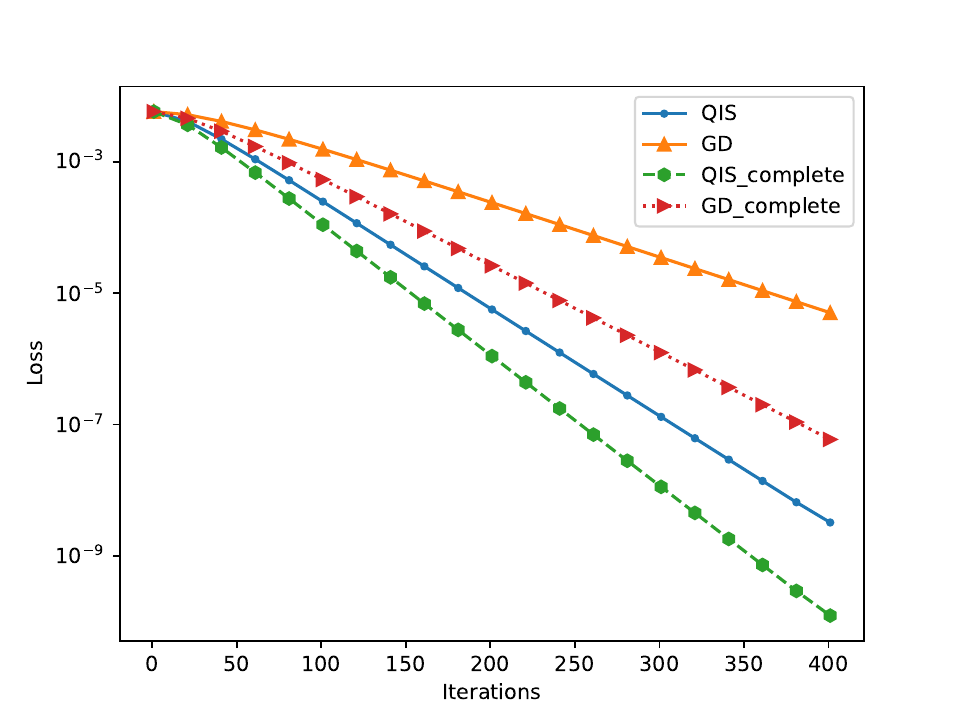}
  }
  \subfigure[Quasi-Newton methods without BB method.]{
    \includegraphics[width=0.6\textwidth]{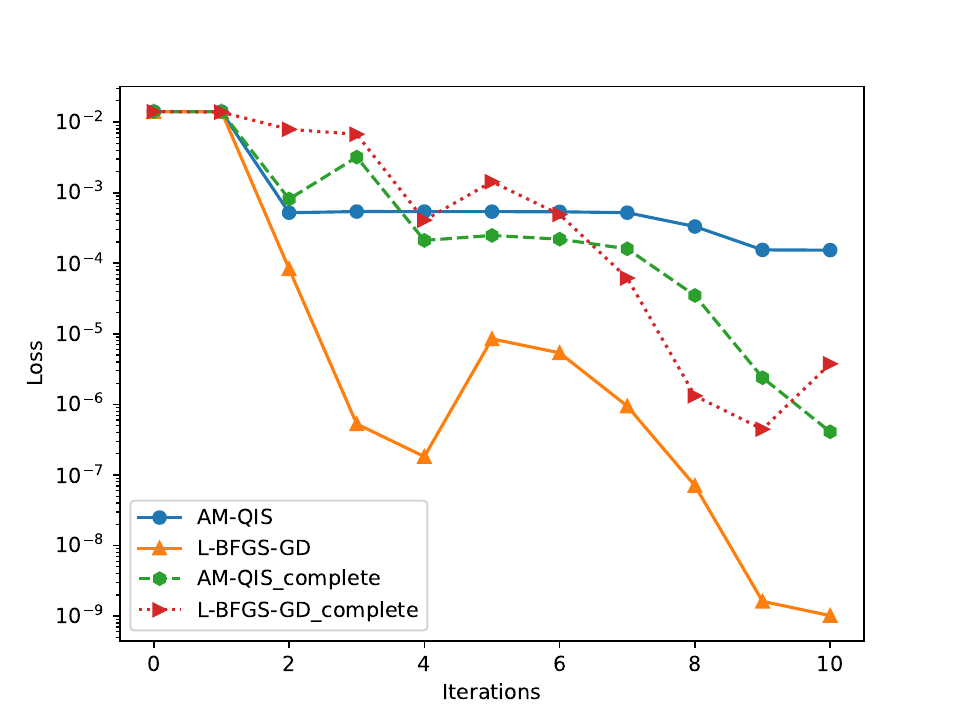}
  }
  \subfigure[Quasi-Newton methods with BB method.]{
    \includegraphics[width=0.6\textwidth]{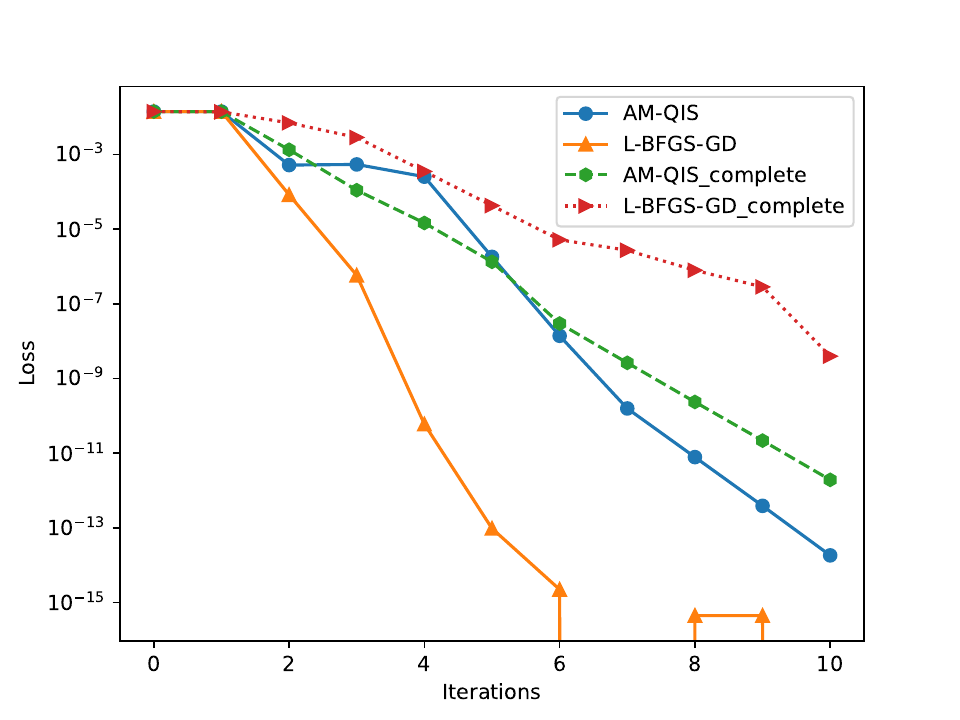}
  }
  \caption{Results of the Ising Hamiltonian for $7$ qubits.}\label{fig:ising}
\end{figure}

The results of the Ising Hamiltonian are shown in \cref{fig:ising} for a
$7$-qubit system with $13$ Hamiltonian terms.
In such cases, the QIS algorithm converges to the optimal solution with error
$10^{-9}$ in roughly $400$ iterations with less loss than the GD algorithm,
while the quasi-Newton methods converge to the optimal solution with similar
error in less than $20$ iterations.

\begin{figure}[ht]
  \centering
  \subfigure[QIS and GD results.]{
    \includegraphics[width=0.6\textwidth]{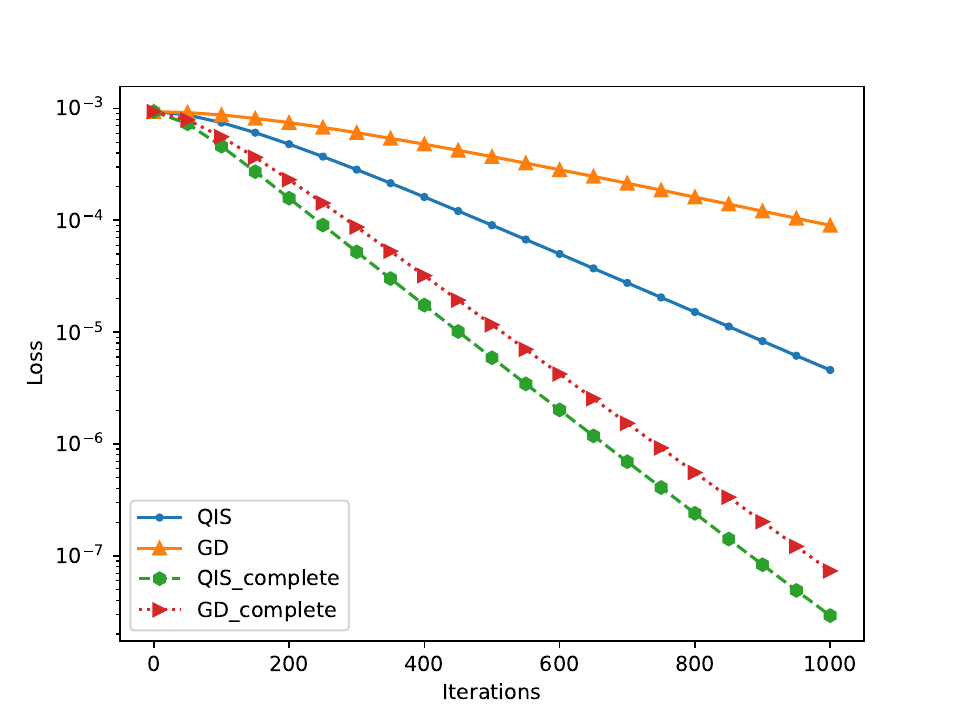}
  }
  \subfigure[Quasi-Newton methods without BB method.]{
    \includegraphics[width=0.6\textwidth]{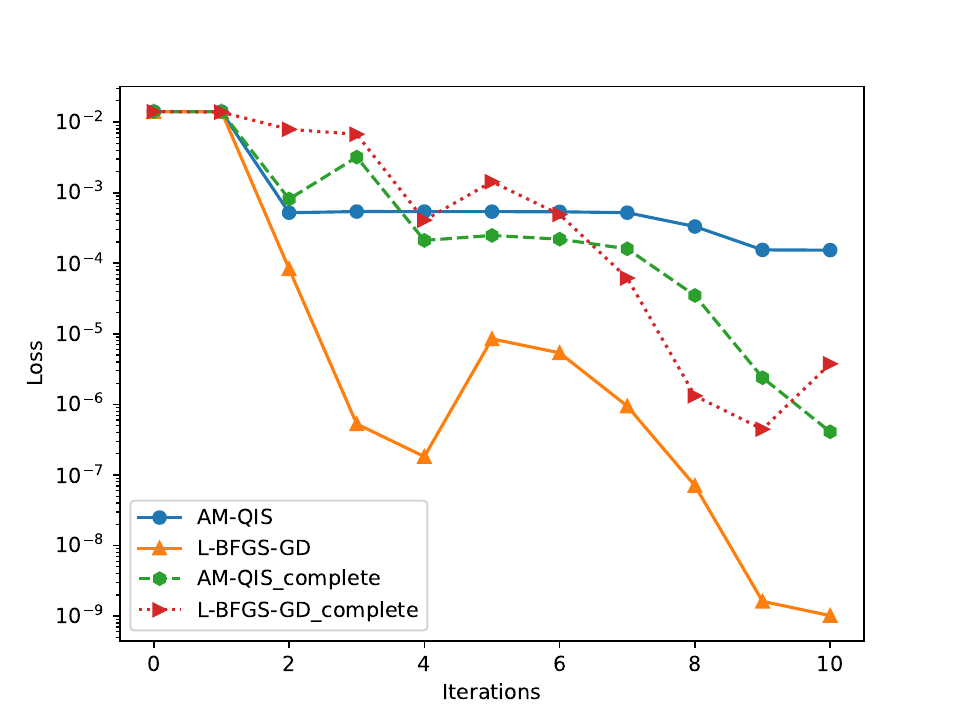}
  }
  \subfigure[Quasi-Newton methods with BB method.]{
    \includegraphics[width=0.6\textwidth]{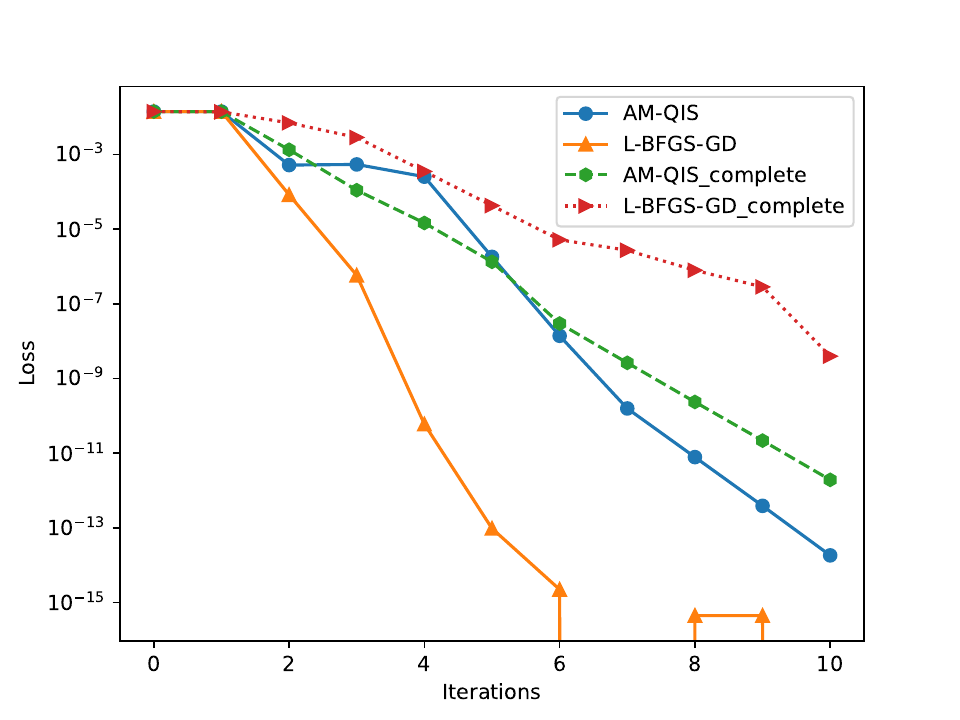}
  }
  \caption{Results of the Transversal1D Hamiltonian for $7$ qubits.}\label{fig:transversal1d}
\end{figure}

The results of the Transversal1D Hamiltonian are shown in
\cref{fig:transversal1d} for a $7$-qubit system with $12$ terms.
In such cases, the QIS algorithm converges to the optimal solution with error
$10^{-7}$ in roughly $1000$ iterations with less loss than the GD algorithm,
while the quasi-Newton methods converge to the optimal solution with similar
error in less than $20$ iterations.

\begin{figure}[ht]
  \centering
  \subfigure[QIS and GD results.]{
    \includegraphics[width=0.6\textwidth]{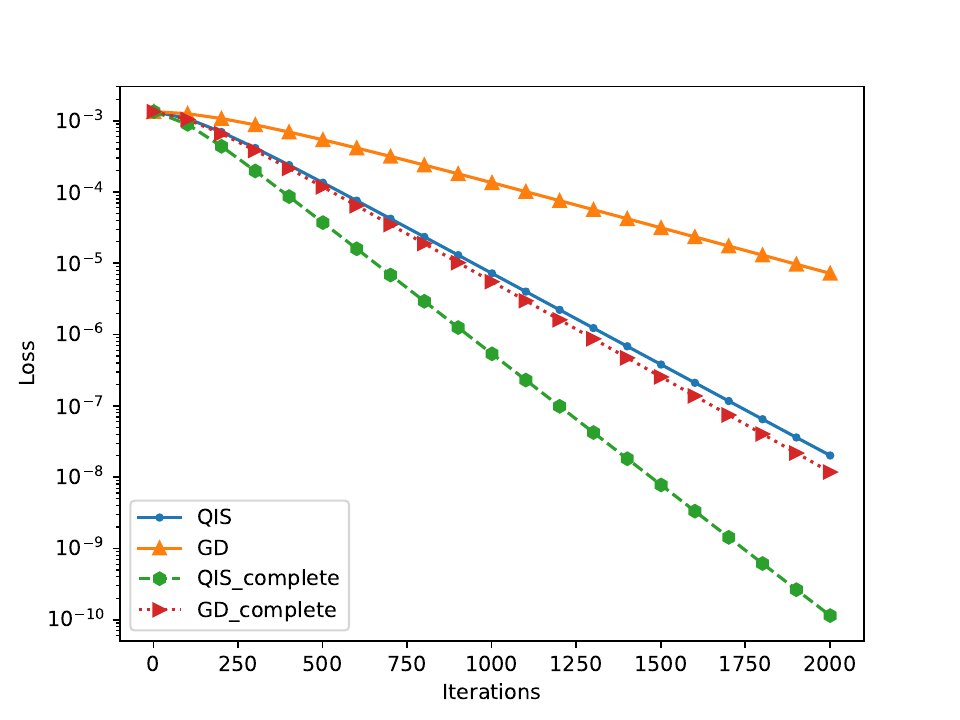}
  }
  \subfigure[Quasi-Newton methods without BB method.]{
    \includegraphics[width=0.6\textwidth]{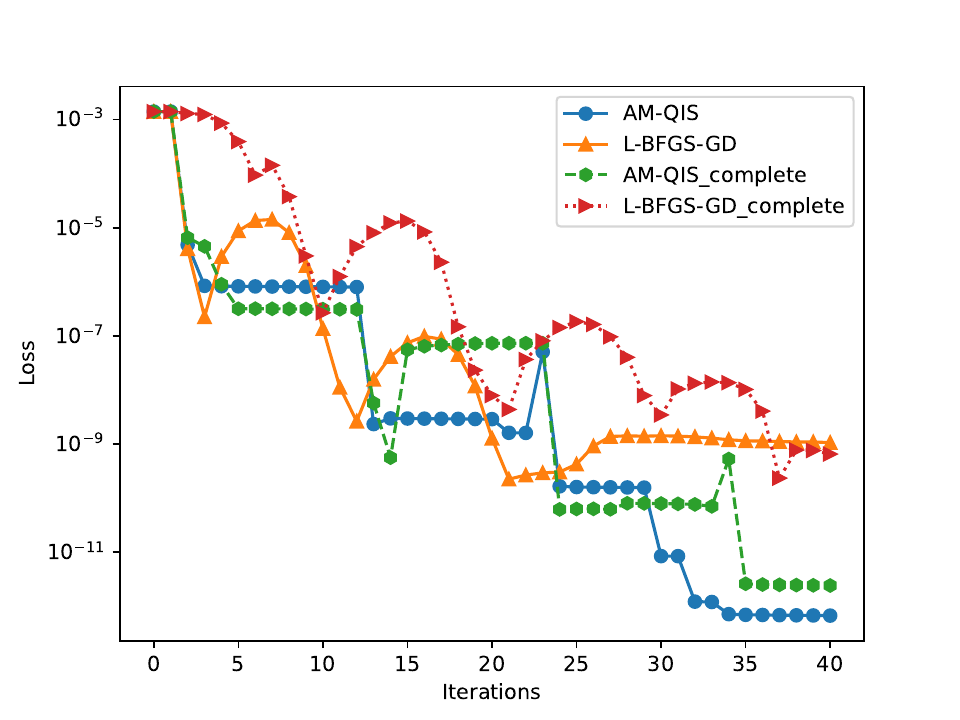}
  }
  \subfigure[Quasi-Newton methods with BB method.]{
    \includegraphics[width=0.6\textwidth]{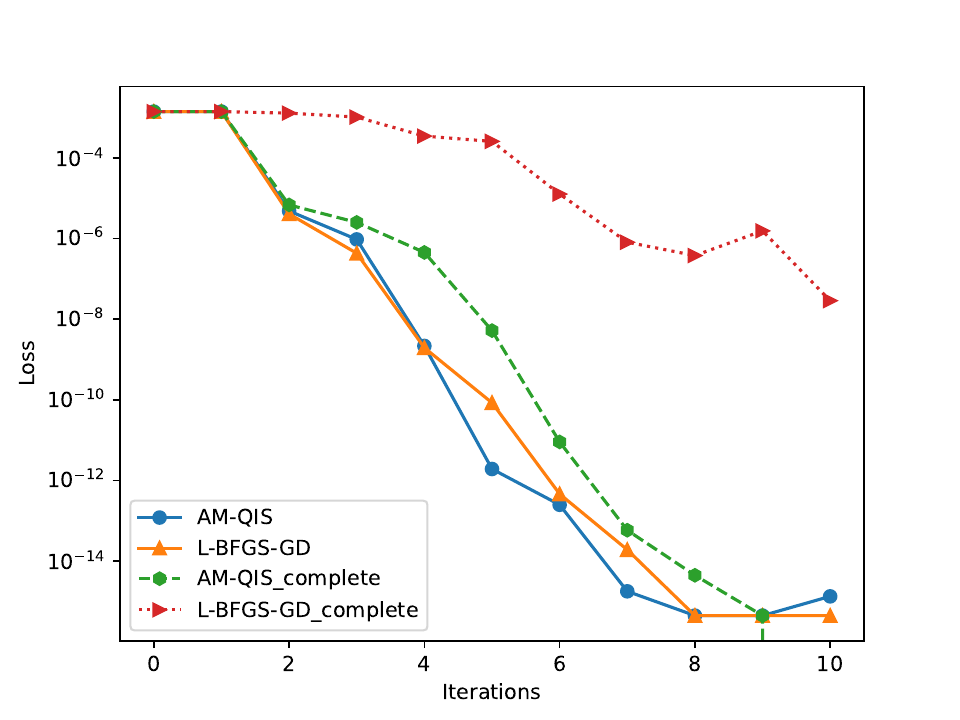}
  }
  \caption{Results of the Local1D Hamiltonian for $7$ qubits.}\label{fig:local1d}
\end{figure}

The results of the Local1D Hamiltonian are shown in \cref{fig:local1d}
for a $7$-qubit system with $96$ terms.
In such cases, the QIS algorithm converges to the optimal solution
with error $10^{-9}$ in roughly $2000$ iterations with less loss than the GD algorithm,
while the quasi-Newton methods converge to the optimal solution with similar error
in less than $40$ iterations.

\subsection{Reproducibility Details}\label{sub-sec:reproducibility}
The numerical experiments were conducted using Python 3.11.5 in Python scripts.
Running the script requires the following packages: numpy and scipy for linear
algebra operations, matplotlib for data visualization, sys for outputting
results in txt format, time for measuring the runtime of the algorithm, and
Qiskit~\citep{Qis23} for generating Hamiltonians.
For reproducibility, we set the random seed to $100$ for QIS and GD, and random
seed array from $1$ to the number of experiments for quasi-Newton accelerations.
The average runtime of the script is less than 6 minutes for systems of five to eight qubits
(of dimensions $32 \times 32$ to $256 \times 256$).

\end{document}